\documentclass{llncs}
\usepackage{epsf}
\usepackage{amssymb}

\begin{document}

\title{Modelling Semantic Association and Conceptual Inheritance for
Semantic Analysis\thanks{This research has been funded by a Marie Curie
Fellowship Grant from the DG of Research, European Commission}}

\author{Pascal Vaillant}

\institute{Humboldt University of Berlin, Chair for Computational Linguistics,\\
           J{\"{a}}gerstra{\ss{}}e 10/11, 10117 {\sc Berlin}, {\sc Germany}\\
           {~}\\
           E-mail: \email{vaillant@compling.hu-berlin.de}}

\maketitle

\begin{abstract}
Allowing users to interact through language borders is an interesting
challenge for information technology.  For the purpose of a computer
assisted language learning system, we have chosen icons for
representing meaning on the input interface, since icons do not depend
on a particular language.  However, a key limitation of this type of
communication is the expression of articulated ideas instead of
isolated concepts.  We propose a method to interpret sequences of
icons as complex messages by reconstructing the relations between
concepts, so as to build conceptual graphs able to represent meaning
and to be used for natural language sentence generation.  This method
is based on an electronic dictionary containing semantic information.
\end{abstract}

\section{Introduction}

There are some contexts in the field of information technology where
the available data is limited to a set of conceptual symbols with no
relations among them.  In applications we have developed, icons are
used on the input interface to represent linguistic concepts for
people with speech disabilities, or for foreign learners of a second
language; in information extraction or indexing applications, sets of
keywords may be given with no higher-level structure whatsoever; the
same situation may occur in a context of cross-linguistic
communication where participants in an online discussion forum are
able to exchange bare concepts through automatic search in electronic
dictionaries, but are not able to master the syntactical structure of
each other's language.

The problem in such contexts is that there is no deterministic way to
compute the semantic relations between concepts; while the meaning of
a structured message precisely resides in the network built from these
relations.  Isolated concepts thus lack the expressive power to convey
ideas: until now, the expression of abstract relations between
concepts still cannot be reached without the use of linguistic
communication.

We have proposed an approach to tackle this
limitation~\cite{vaillant:97}: a method to interpret sequences of
isolated concepts by modelling the use of ``natural'' semantic
knowledge is implemented.  This allows to build knowledge networks
from icons as is usually done from text.  A first application,
developed for a major electronics firm, had aimed at proposing
speech-impaired people an iconic aided communication software.  We are
now working at improving the theory in order to implement it in the
field of computer assisted language learning.  Here we present new
formalisms to model lexical meaning and associative semantic
processes, including representation of conceptual inheritance,
which have been developed for the latter application.

\section{Description of the problem}

Assigning a {\em signification} to a sequence of information items
implies building conceptual relations between them. Human linguistic
competence consists in manipulating these dependency relations: when
we say that ``the cat drinks the milk'', for example, we perceive that
there are well-defined conceptual connections between `cat', `drink',
and `milk'---that `cat' and `milk' play given roles in a given
process.  Linguistic theories have been developed specifically to give
account of these phenomena \cite{tesniere:88,melcuk:88}, and several
symbolic formalisms in AI \cite{schank:75,sowa:84} reflect the same
approach. Computationally speaking, `cat', `drink' and `milk' are:
without relations, a set of keywords; with relations, a structured
information pattern. This has important consequences e.g. in text
filtering and information retrieval.

Human natural language reflects these conceptual relations in its
messages through a series of linguistic clues. These clues, depending
on the particular languages, can consist mainly in word ordering in
sentence patterns (``syntactical'' clues, e.g. in English, Chinese, or
Creole), in word inflection or suffixation (``morphological'' clues,
e.g. in Russian, Turkish, or Latin), or in a given blend of both
(e.g. in German). {\em Parsers} are systems designed to analyze
natural language input, on the base of such clues, and to yield a
representation of its informational contents.

In the context of language learning, where icons have to be used to
convey complex meanings, the problem is that morphological clues are
of course not available, when at the same time we cannot rely on a
precise sentence pattern (there is no ``universal icon grammar'', and
if we were addressing perfectly functional speakers of a given
language, with its precise set of grammar rules, we wouldn't be using
icons).

Practically, this means that, if we want to use icons as an input for
computer communication, we cannot rely on a parser based on phrase
structure grammar (``CFG''-style) to build the conceptual relations of
the intended message. We should have to use a parser based on
dependency computing, such as some which have been written to
cope with variable-word-order languages \cite{covington:90}. However,
since no morphological clue is available either to tell that an
icon is accusative or dative, we have to rely on semantic knowledge
to guide role assignment. In other words, an icon parser has to
know that drinking is something generally done by living beings
and involving liquid objects.

\section{Modelling Meaning}

The first step is then to encode the semantic information representing
this type of natural world knowledge. For this purpose, we develop an
icon lexicon where the possible semantic relations are specified by
feature structures among which unification can take place. However,
the feature structures do not have a syntactic meaning here, like
e.g. in HPSG, but a natural language semantics meaning: Instead of
formal grammatical features, it is specified which ``natural
properties'' the different icons should have, and how they can combine
with the others.

\subsection{Intrinsic vs. extrinsic features}

Every icon in the lexicon has a certain number of {\em intrinsic}
attributes, defining its fundamental meaning elements. Going back to
our example, `cat' has the features {\em animal}\,, {\em living}\,,
while `milk' has the features {\em liquid}\,, {\em food}\,.

In natural language semantics, some pair of concepts are defined in
opposition to each other; for the sake of modelling simplicity, we
define these pairs as couples of features sharing the same attribute
but with an opposite value. This modelling choice leads to define the
basic feature, or intrinsic feature, as a pair
$\langle{}a,v\rangle{}$, where the attribute $a$ is a symbol, and the
value $v$ is $+1$ or $-1$.

Yet intrinsic features are not enough to build up relations: we need
at least some first-order semantics to allow predication. Hence a
restricted set of icons, the predicative icons (roughly corresponding
to natural language verbs and adjectives), also have sets of {\em
extrinsic} (or selectional) features, that determine which other
concepts they may incorporate as actants. These extrinsic features
specify for example which properties are ``expected'' from the agent
or the object of an action, or to which categories of concepts a
particular adjective may be attributed: in our example, `drink' would
have the features {\em agent(animal)} and {\em object(beverage)}.

This could lead to define the extrinsic feature as a pair
$\langle{}c,${\em ef}\,$\rangle{}$, where the case $c$ is a symbol, and the
expected feature {\em ef}\, is an intrinsic feature as defined above,
i.e. as being of the form $\langle{}c,\langle{}a,v\rangle{}\rangle{}$,
where case $c$ and attribute $a$ are symbols, and value $v$ is $+1$ or
$-1$.

However, with such a definition, the selectional effect of an
extrinsic feature can only be {\em compelling} (the attribute is
present with a value of $+1$), {\em blocking} (the attribute is
present with a value of $-1$), or null (the attribute is absent).  Yet
natural semantics involves the ability to represent gradation: in
natural language for instance, a given association between words may
be {\em expected}, but it does not completely block the possibility
that another one be realized.

So, we decide to define the extrinsic feature as
$\langle{}c,\langle{}a,v\rangle{}\rangle{}$, where $c$ (the case) is a
symbol, $a$ (the attribute) is a symbol, and $v \in{} \bbbr{}$. This
way of modelling allows to tune the value $v$ in order to make a
semantic association more or less compelling.

The extrinsic features contain all the information about the potential
case relations that may occur in the icon language. Considering a
given predicative icon, its valency frame, or case frame, is strictly
equivalent to the set of its extrinsic features factorized by case.
Considering the whole lexicon, the case system is defined by the set
of all cases appearing in any extrinsic feature of any icon.

\subsection{Feature inheritance}

There are obvious advantages of including a representation of
inheritance in the lexicon, such as: saving representation space
(`dog', `cat', and `hamster' only have a few specific features
represented separately, the rest is stored under `pet'); providing a
measure of semantic distance between concepts (how many ``common
ancestors'' do they have, and at which level?)

However, since natural concepts may be grouped in overlapping
categories, there can be no unique tree-like hierarchy covering the
whole lexicon. For this reason, a mechanism of multiple inheritance
has been developed.

The multiple inheritance model allows a single concept to inherit
thematic features from a thematic group, as well as structural
features from an abstract superconcept spanning different subgroups of
the thematic hierarchy (like for instance the superconcept `action',
which passes the extrinsic feature {\em agent(animal)} on to all
specific concepts which inherit from it). Intrinsic features as well
as extrinsic features may be inherited, and passed on to more specific
subconcepts.

The well-known theoretical problem of multiple inheritance, namely the
possibility that a concept inherit contradictory features from two
separate branches of the inheritance graph, is not an actual problem
in the context of a model for natural meaning. In fact, natural
categories are not logical categories, and it is actually normal that
contradictions may arise. If they do, they are meaningful and should
not be ``solved''. Specifically, in the analysis application described
below, feature values are added, so if an attribute appears once with
a positive value, and once with a negative one, it counts as if its
value were zero.

It is important to note that concept labels may be used as attributes
in semantic features, like when we want to specify that the object of
`drink' has to be a `beverage'. This means that we do not postulate
any ontological difference between a feature and a concept. As a
matter of fact, studies in natural language semantics, for instance,
always represent features by using words: features simply are more
primitive concepts than the concepts studied. So, when we say that
concept {\em a} inherits from concept {\em b} (is a subconcept of
concept {\em b}), we mean exactly the same thing as when we say that
concept {\em a} has the feature {\em b}, and there is a unique formal
representation for all this, like in the example below:

\begin{center}
\epsfxsize=120mm
\mbox{\epsffile{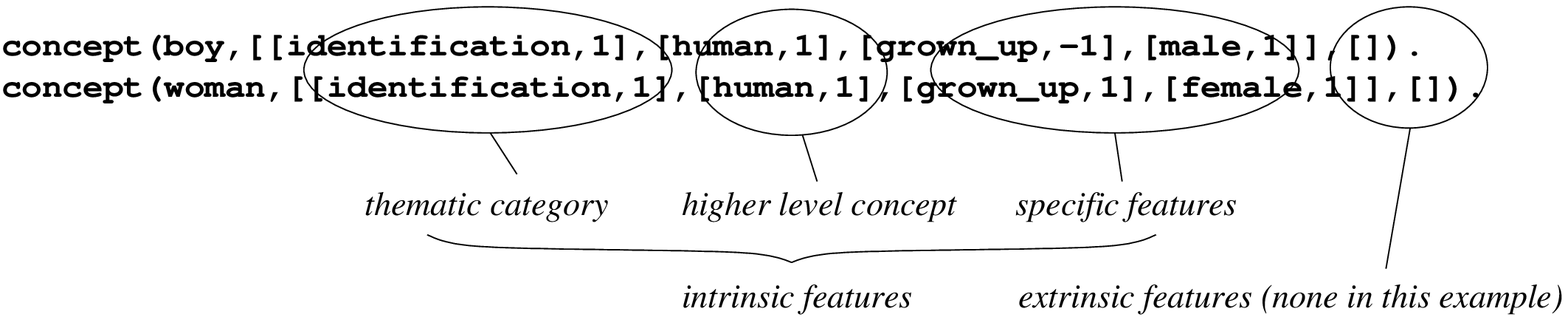}}
\end{center}

An important practical consequence of this is that we can talk of
feature inheritance: this will be used in the analysis process.

\section{The semantic analysis method}

The icon parser we propose performs semantic analysis of input
sequences of icons by the use of an algorithm based on
best-unification: when an icon in the input sequence has a
``predicative'' structure (it may become the head of at least one
dependency relation to another node, labeled ``actant''), the other
icons around it are checked for compatibility.

Semantic compatibility is then measured as a unification score between
two sets of feature structures: the intrinsic semantic features of the
candidate actant:

${\cal IF} = \{ \langle{}a_{11},v_{11}\rangle{}, \langle{}a_{12},v_{12}\rangle{},
\dots{}, \langle{}a_{1m},v_{1m}\rangle{} \}\enspace{},$

and the extrinsic semantic features of the predicative icon attached
to the semantic role considered, the case $c$\,:

${\cal SF} = \{ \langle{}a_{21},v_{21}\rangle{}, \langle{}a_{22},v_{22}\rangle{},
\dots{}, \langle{}a_{2n},v_{2n}\rangle{} \}\enspace{},$

(where $\langle{}c,\langle{}a_{21},v_{21}\rangle{}\rangle{},
\langle{}c,\langle{}a_{22},v_{22}\rangle{}\rangle{}, \dots{},
\langle{}c,\langle{}a_{2n},v_{2n}\rangle{}\rangle{}$ are
extrinsic features of the predicate).

The basic idea is to define compatibility as the sum of matchings in
the two sets of attribute-value pairs, in ratio to the number of
features being compared to. Note that semantic compatibility is not a
symmetric norm: it has to measure how good the candidate actant
(i.e. the set ${\cal IF}$) fits to the expectations of a given
predicative concept in respect to its case $c$ (i.e. to the set ${\cal
SF}$). Hence there is a {\em filtering} set (${\cal SF}$) and a {\em
filtered} set (${\cal IF}$). The asymmetry shows itself in the
following definition of the compatibility function, in that the
denominator is the cardinal of ${\cal SF}$, not of ${\cal IF}$:

\begin{eqnarray*}
{\cal C}({\cal IF},{\cal SF}) & = & {\cal C}(\{\langle{}a_{11},v_{11}\rangle{},\dots{},\langle{}a_{1m},v_{1m}\rangle{}\},\{\langle{}a_{21},v_{21}\rangle{},\dots{},\langle{}a_{2n},v_{2n}\rangle{}\}) \\
 & = & \frac{\sum_{j \in [1,n]}{\sum_{i \in [1,m]}{f(\langle{}a_{1i},v_{1i}\rangle{},\langle{}a_{2j},v_{2j}\rangle{})}}}{n}\enspace{},
\end{eqnarray*}

where $f$ is a matching function defined on pairs of individual
features, not on pairs of sets of features.

Now the compatibility function $f$ has to be defined at the level of
the features themselves so as to take into account the inheritance
phenomena. So we define
$f(\langle{}a_{1},v_{1}\rangle{},\langle{}a_{2},v_{2}\rangle{})$
(where $\langle{}a_{1},v_{1}\rangle{}$ is the intrinsic $[$filtered$]$
feature, and $\langle{}a_{2},v_{2}\rangle{}$ the extrinsic
$[$filtering$]$ feature), as following:

\begin{itemize}

\item[1 --] If the two attributes are the same ($a_{1} = a_{2} = a$\,):

\( f(\langle{}a,v_{1}\rangle{},\langle{}a,v_{2}\rangle{}) = v_{1}.v_{2} \)\enspace{};

\item[2 --] if $a_{1} \Rightarrow{} a_{2}$ ($a_{1}$ includes $a_{2}$ in its
signification, i.e. $a_{1}$ is a subtype of $a_{2}$\,):

\begin{tabular}{rclcl}
- & \makebox[1ex]{} & if $v_{1} < 0$\,, & \makebox[2ex]{} & $f(\langle{}a_{1},v_{1}\rangle{},\langle{}a_{2},v_{2}\rangle{}) = 0$\,, \\
- & \makebox[1ex]{} & if $v_{1} \geq{} 0$\,, & \makebox[2ex]{} & $f(\langle{}a_{1},v_{1}\rangle{},\langle{}a_{2},v_{2}\rangle{}) = v_{1}.v_{2}$\,; \\
\end{tabular}

\item[3 --] if $a_{1} \Rightarrow{} \overline{a_{2}}$ ($a_{1}$ includes a feature
$a'_{2}$ in its signification, such that $a'_{2}$ is contradictory
with $a_{2}$\,):

\begin{tabular}{rclcl}
- & \makebox[1ex]{} & if $v_{1} < 0$\,, & \makebox[2ex]{} & $f(\langle{}a_{1},v_{1}\rangle{},\langle{}a_{2},v_{2}\rangle{}) = 0$\,, \\
- & \makebox[1ex]{} & if $v_{1} \geq{} 0$\,, & \makebox[2ex]{} & $f(\langle{}a_{1},v_{1}\rangle{},\langle{}a_{2},v_{2}\rangle{}) = -v_{1}.v_{2}$\,; \\
\end{tabular}

\item[4 --] if $a_{1} \neq a_{2}$\,, and $a_{1} \nRightarrow{} a_{2}$\,, and $a_{1} \nRightarrow{} \overline{a_{2}}$\,, then:

\begin{tabular}{rcl}
- & \makebox[1ex]{} & either $a_{2}$ is a primitive feature ($\nexists{} x\enspace{}|\enspace{}a_{2} \Rightarrow{} x$\,), in which case: \\
 & \makebox[1ex]{} & \makebox[1ex]{}$f(\langle{}a_{1},v_{1}\rangle{},\langle{}a_{2},v_{2}\rangle{}) = 0$\,, \\
- & \makebox[1ex]{} & or $a_{2}$ is decomposable in more primitive features; and then:\\
 & \makebox[1ex]{} & let $\{a_{21},a_{22},\dots{},a_{2k}\}$ the set of features implied by $a_{2}$ \\
 & \makebox[1ex]{} & \makebox[1ex]{}($a_{2} \Rightarrow{} a_{2j}\mbox{ for }j \in [1,k]$\,) \\
 & \makebox[1ex]{} & then \\
 & \makebox[1ex]{} & \makebox[1ex]{}$f(\langle{}a_{1},v_{1}\rangle{},\langle{}a_{2},v_{2}\rangle{}) =$ \\
 & \makebox[1ex]{} & \makebox[3ex]{}${\cal C}(\{\langle{}a_{1},v_{1}\rangle{}\},\{\langle{}a_{21},v_{2}\rangle{},\langle{}a_{22},v_{2}\rangle{},\dots{},\langle{}a_{2k},v_{2}\rangle{},\langle{}\mbox{\tt dummy\_symbol},v_{2}\rangle{}\})$\enspace{}.
\end{tabular}

\end{itemize}

Let us explain and illustrate this definition by simple examples.
Suppose we want to test whether some icon possessing the feature
{\em dog}
($\langle{}\mbox{\tt dog},1\rangle{}$)
is a good candidate for being the agent of the verb `bark'; `bark'
having an extrinsic feature
{\em agent(dog)}
($\langle{}\mbox{\tt agent},\langle{}\mbox{\tt dog},1\rangle{}\rangle{}$\,).
We will then be trying to evaluate
$f(\langle{}\mbox{\tt dog},1\rangle{},\langle{}\mbox{\tt dog},1\rangle{})$\,.
This is the case 1, and the result will be 1.  If we had tried to
match this same icon to a verb whose agent {\em should
not} be a dog
($\langle{}\mbox{\tt agent},\langle{}\mbox{\tt dog},-1\rangle{}\rangle{}$\,),
the result would of course have been $-1$.

Now suppose we want to match {\em dog} to a verb which only expects
its agent to be an animal. We will have to evaluate
$f(\langle{}\mbox{\tt dog},1\rangle{},\langle{}\mbox{\tt animal},1\rangle{})$\,.
{\em dog} being a subtype of {\em animal}\,, we have
$\mbox{\tt dog} \Rightarrow{} \mbox{\tt animal}$\,, so we are
in the case 2, and the result is 1 (a dog fulfills entirely the
expectation of being an animal).

If on the other hand we wanted to match some concept of which
we know it is {\em not} a dog, because it has the feature
$\langle{}\mbox{\tt dog},-1\rangle{}$\,, to the semantic role
where an animal is expected, we could obviously draw no conclusion
from the only fact that it is not a dog. Not being a dog does
not imply not being an animal. This is why in this particular
subcase of case 2, the result is 0.

Now if we want to match {\em dog} to some semantic role where an
{\em object} is expected, we find that
$\mbox{\tt dog} \Rightarrow{} \mbox{\tt living\_being}$\,,
and {\em object} and {\em living being} being mutually exclusive,
we are in the case 3 and find the value $-1$.

Like in case 2, there is a subcase of case 3 where the result is $0$
because no conclusion can be drawn (e.g. we can not deduce from
something not being a dog that it is a non living object).

Finally let us suppose that we want to match some animal which is not
a dog to the agent role of `bark', which expects {\em dog}\,. The
candidate concept does not possess the feature
$\langle{}\mbox{\tt dog},1\rangle{}$
but it possesses the feature
$\langle{}\mbox{\tt animal},1\rangle{}$\,.
It would be inappropriate, in this case, that this concept should have
no better score than any other: being an animal, it is semantically
``closer'' to {\em dog} than an inanimate object, for example, would
be (this is what allows, in natural language semantics, sentences
like ``the police superintendent barks'' \cite{greimas:66}).

This is why, in this case, we break up {\em dog} into more primitive
components and recursively call the function ${\cal C}$ (compatibility
on sets of features), so that
$\langle{}\mbox{\tt animal},1\rangle{}$
will eventually meet
$\langle{}\mbox{\tt animal},1\rangle{}$\,,
and will yield a positive, though fractional, result.

A dummy feature is added so that the compatibility value loose a small
proportion of itself in this operation of breaking up, by incrementing the
denominator.

Note that the recursivity (${\cal C}$ is based on $f$ and $f$
is---partially---based on ${\cal C}$\,) is not infinite, since
the decomposition always falls back on primitive features: there
is no infinite loop. This is guaranteed, not by the definition
of the functions themselves, but by the fact that the inheritance
graph is a direct graph.

Globally, for every predicate in the actual input sequence, the analysis
process seeks to assign the best actant for every possible role of the
predicate's immanent conceptual structure. The absolute compatibility
between the predicate and the actant, defined in the sense of the
function ${\cal C}$ described above, is weighted by a function valued
between 0 and 1 and decreasing with the actual distance between the
two icons in the sequence.

The result yielded by the semantic parser is the graph that maximizes
the sum of the compatibilities of all its dependency relations. It
constitutes, with no particular contextual expectations, and given the
state of world knowledge stored in the iconic database in the form of
semantic features, the ``best'' interpretation of the users' input.

\section{Application and Evaluation}

A primitive version of the semantic analysis algorithm has been
implemented in 1996 for rehabilitation purposes, within a French
electronics firm (Thomson-CSF), in the frame of a software
communication tool for speech-impaired people \cite{vaillant:97}. The
evaluation led to acceptable performance in analysis accuracy (80.5 \%
of the sequences correctly analyzed on a benchmark of 200
samples). However the acceptance level by the user remained low, due
to a strongly time consuming recursive algorithm (the complexity and
time grew in a $O(n.e^{n})$ relation to the size of the input).

\begin{figure}
\begin{center}
\epsfxsize=110mm
\mbox{\epsffile{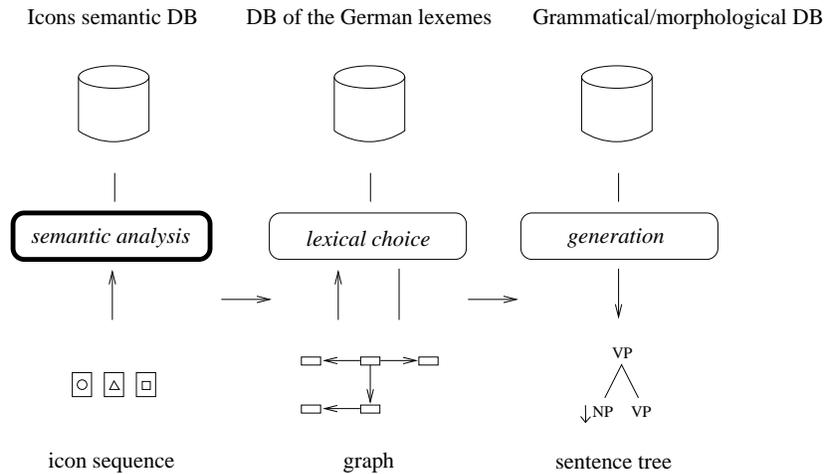}}
\end{center}
\caption[system-structure]{Structure of the CALL system}
\label{system-structure}
\end{figure}

An application to the field of CALL (Computer Assisted Language
Learning) is currently being developed at the Humboldt University of
Berlin.  The application prototype aims at allowing learners of German
as a second language to practice communication in that language at
home or in tutorial classes. The users first tell the computer what
they intend to express by pointing to icons. The system interprets
these icons semantically, and proposes a choice of rated formulations
(1) in the form of conceptual graphs, and (2) as full German
sentences. The users are then allowed to ``play'' with the graph to
discover how to express variations or refinements, in particular
concerning nuances in verbs like expressed in Kunze's theory of verb
fields \cite{kunze:93}. This application is made possible by mapping
the results of the semantic analysis into a lexical database of the
German language developed by the Chair for Computational Linguistics
at the Humboldt University (Fig. \ref{system-structure}).

The implementation principles have been renewed in this application,
so as to develop a form of parser storing its intermediate results
(inspired by ``chart parsers'' for CFG grammars). This allows
considerably less backtracking, and hence a big gain in computational
complexity (now measured in $O(n^{2})$), and removes one of the major
impediments of the method.


\end{document}